# Cooperative Collision Avoidance in a Connected Vehicle Environment

**Sukru Yaren Gelbal, Sheng Zhu, Gokul Arvind Anantharaman, Bilin Aksun Guvenc, and Levent Guvenc**

Automated Driving Lab, Ohio State University

## Abstract

Connected vehicle (CV) technology is among the most heavily researched areas in both the academia and industry. The vehicle to vehicle (V2V), vehicle to infrastructure (V2I) and vehicle to pedestrian (V2P) communication capabilities enable critical situational awareness. In some cases, these vehicle communication safety capabilities can overcome the shortcomings of other sensor safety capabilities because of external conditions such as 'No Line of Sight' (NLOS) or very harsh weather conditions. Connected vehicles will help cities and states reduce traffic congestion, improve fuel efficiency and improve the safety of the vehicles and pedestrians. On the road, cars will be able to communicate with one another, automatically transmitting data such as speed, position, and direction, and send alerts to each other if a crash seems imminent. The main focus of this paper is the implementation of Cooperative Collision Avoidance (CCA) for connected vehicles. It leverages the Vehicle to Everything (V2X) communication technology to create a real-time implementable collision avoidance algorithm along with decision-making for a vehicle that communicates with other vehicles. Four distinct collision risk environments are simulated on a cost effective Connected Autonomous Vehicle (CAV) Hardware in the Loop (HIL) simulator to test the overall algorithm in real-time with real electronic control and communication hardware.

## Introduction

Cooperative vehicle applications based on V2V communications are promising to significantly reduce accidents, alleviate traffic congestion, and improve energy consumption [1]. With the advances in the wireless communication technology in recent years, real-world prototype implementations of CV applications have started to take place by making use of the real-time shared information about traffic, vehicle states and the road environment [2-3]. Of these applications, the ones related to emergency safety and collision avoidance draw the most attention for increasing road safety of passengers and other road users. The U.S. Department of Transportation (DOT) has estimated that vehicle-to-vehicle (V2V) systems could address around 79% of all vehicle crashes in the United States [4], potentially saving thousands of lives and billions of dollars. Several scenarios of high crash risk [5] have been noticed in the report due to a lack of driver information. With the aid of V2V communication, there are plenty of reported research efforts on collision prevention in these specific scenarios including intersection [6-7], and sudden stop of the lead vehicle [8-9]. In these papers, a good prediction of other vehicles' behavior based on

V2V information and corresponding decision-making are essential to avoid potential collisions.

However, cooperative collision avoidance applications are generally time-sensitive and require immediate decision-making and following action implementations. This demands that the CCA algorithms should also be simple enough with the capability of real-time implementation. To address this issue, this paper considers several scenarios of high crash risk and addresses the corresponding decision-making algorithm aided by V2V communication. To validate the real-time performance of the developed CCA applications, hardware-in-the-loop (HIL) tests were conducted with real DSRC on-board-unit (OBU) modems for inter-vehicular communication.

This paper is organized as follows. Section II introduces the path following controller which is the elastic band method used to form a collision-free path along with the decision-making module. Section III describes the setup of hardware-in-the-loop simulator to simulate V2V scenarios with DSRC communication. This is followed by real-time HIL simulation results and discussion of four selected scenarios of interest in Section IV. The paper ends with conclusions in the last section.

## Methods

### Path Following

For path following, map generation and error calculation are two main parts. The path is generated using GPS waypoints which are recorded previously with manual driving or obtained from an online map. The route is then divided into segments which are smaller parts of the road, each containing an equal number of points. Third order polynomial curves are fitted for each of these segments using Eq. 1 and 2.

$$X_i(\lambda) = a_{xi}\lambda^3 + b_{xi}\lambda^2 + c_{xi}\lambda + d_{xi} \tag{1}$$

$$Y_i(\lambda) = a_{yi}\lambda^3 + b_{yi}\lambda^2 + c_{yi}\lambda + d_{yi} \tag{2}$$

The polynomial coefficients $a, b, c, d$ in equations (1) and (2) are generated offline and they contain all the path information. For the second part, these generated coefficients are fed into the error calculation. Lateral deviation error and yaw angle error are calculated with respect to this path's curvature information around the vehicle position with a predefined preview distance. Steering is controlled according to this error. The reader is referred to several previous studies like [10-11] for more detailed information about the path determination and following method that is used here.

## Collision Avoidance with Elastic Band

While doing path following, in case of an obstacle appearing on the path or near the path, the vehicle switches to collision avoidance mode. Path points close to the obstacle are moved away with forces according to the position of the obstacle. Therefore, a modified path is generated online, while the vehicle is still doing the path following. Then, instead of the pre-defined path, these modified points are followed in order to maneuver around the obstacle. Figure 1 illustrates the forces acting on the modified path points which were originally crossing the obstacle. Elastic band method can be applied naturally to a road vehicle path following task where there is a pre-defined path that lies between the lanes with collision avoidance maneuvering being limited at most to emergency (sudden) lane changes. Another main reason is the real-time computation capability. Some of the advantages and disadvantages for different path planning algorithms are shown in Table 1.

Table 1. Advantages and disadvantages for some path planning methods

| Method | Advantages | Disadvantages |
|---|---|---|
| Motion Primitives | Very simple and very fast. | Route is neither optimal nor flexible. Behavior can be unnatural for a car. |
| Rapidly-Exploring Random Tree | Easy to implement. Suitable for holonomic kinematics. | Multiple random searches are relatively time consuming. Can cause unnatural behavior for a car due to randomness. |
| Model Predictive Control | On-line path generation and modification are combined. | With constraints and weights, calculation of optimization is relatively slow. |

Elastic band method in this paper is borrowed from our earlier work [12]. In this method, the original path determined by path planning is divided into nodes (knots) connected by elastic strings that hold the path together using internal forces $F_{int}$. External forces $F_e$ act on the band representing the path when a vehicle (obstacle) crosses the original path. The external forces bend the path like an elastic band such that a collision free deformed path is obtained in natural manner. Figure 1 shows the change in internal forces as the path deforms under the action of external forces generated by a vehicle in a collision path. When the system dynamics are ignored, variation of each internal force can be defined as,

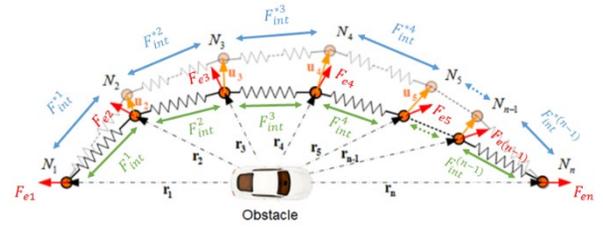

Figure 1. Modification of the points on the path.

$$F_{int}^{*i} - F_{int}^i = k_s(u_{i+1} - u_i) \tag{3}$$

In the equation above, $F_{int}^{*i}$ is final and $F_{int}^i$ is the initial internal force between the $i^{th}$ and $(i+1)^{th}$ elastic band node. $u_i$ is the displacement of the $i^{th}$ knot and $k_s$ is the spring constant. Moreover, the external force $F_{ei}$ acting on the $i^{th}$ knot can be expressed as,

$$F_{ei} = -[k_s(u_{i+1} - u_i) + k_s(u_{i-1} - u_i)] \tag{4}$$

We can write the external forces $F_{ei}$ with their $x$ and $y$ components in matrix form as,

$$F_{ex,ey} = k_s K u_{x,y} \tag{5}$$

where

$$F_{ex,ey} = \begin{bmatrix} F_{ex1} & F_{ey1} \\ F_{ex2} & F_{ey2} \\ \vdots & \vdots \\ F_{exn} & F_{eyn} \end{bmatrix} \quad u = \begin{bmatrix} u_{x1} & u_{y1} \\ u_{x2} & u_{y2} \\ \vdots & \vdots \\ u_{xn} & u_{yn} \end{bmatrix} \tag{6}$$

and

$$K = \begin{bmatrix} -1 & 2 & -1 & 0 & 0 & \dots & 0 \\ 0 & -1 & 2 & -1 & 0 & \dots & 0 \\ \dots & \ddots & \ddots & \ddots & \ddots & \vdots & \vdots \\ 0 & 0 & \dots & \dots & -1 & 2 & -1 \end{bmatrix} \tag{7}$$

External forces $F_{ex,ey}$ acting on each node can be calculated using,

$$F_{ex,ey} = \begin{cases} -ke(\|r_{x,y}\| - r_0)\dfrac{r_{x,y}}{\|r_{x,y}\|}, \|r_{x,y}\| \le r_0 \\ 0, \|r_{x,y}\| > r_0 \end{cases} \tag{8}$$

where $r_{x,y}$, shown as $r_i$ in Figure 1., is the position vector between each node and the obstacle. $r_0$ is the threshold distance and $k_e$ is the stiffness constant value for the external force.

If equations (5) and (8) are combined, the displacement of the nodes can be determined as,

$$u_{x,y} = (\tfrac{1}{k_s})K^{-1}F_{ex,ey} \tag{9}$$

The modified path should start and end on the initial undeformed path. In order to achieve this, the first and the last node should not be moved. Therefore, displacement of the first node should be $u_{x1,y1} = [0,0]$ and displacement of the last node should be $u_{xn,yn} = [0,0]$. Lastly, the final position of the nodes $P_{x,y}^{def}$ on the deformed path can be obtained as

$$P_{x,y}^{def} = P_{x,y} + u_{x,y} \qquad (10)$$

where $P_{x,y}$ is the starting position of the node and $u_{x,y}$ is the displacement of the node.

### Decision-Making for Lane Change Scenarios

While the ego vehicle is avoiding or maneuvering around the other vehicles on its desired path, it needs to consider the vehicles on the adjacent lane to prevent any accidents. This adjacent vehicle can also be approaching from either the same or the opposite direction. An example of the adjacent vehicle approaching from the opposite direction is shown in Figure 2. This illustration also shows information about the terms in equation (11).

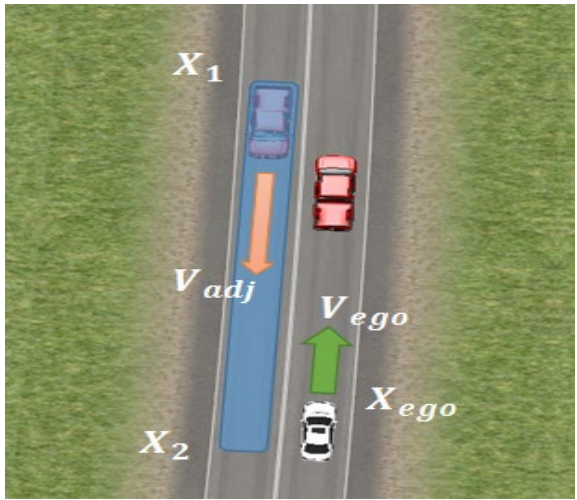

Figure 2. Decision-Making criteria.

$$X_2 = X_1 + (V_{adj} - V_{ego})t_{maneuver} + x_{safety} \qquad (11)$$

The adjacent vehicles create a danger zone according to equation (11). $X_2$ is the end point of the danger zone which is calculated and $X_1$ is start position of the adjacent vehicle body, $V_{adj}$ is the velocity of the adjacent vehicle, $V_{ego}$ is velocity of the ego vehicle, $t_{maneuver}$ is the time it takes to complete the maneuver and $x_{safety}$ is the additional safety distance for extra tolerance. After this calculation is done in very step, if the ego vehicle is within the danger zone in the longitudinal direction, it does not maneuver. Instead, it adapts to the speed of the preceding vehicle while waiting for the danger zone to pass. After the adjacent vehicle is out of the danger zone, the ego vehicle executes the maneuver collision avoidance maneuver with the preceding vehicle. This maneuver is in essence similar to an automatically executed double lane change maneuver.

### HIL Setup

In this study, a HIL setup is used to evaluate the effectiveness of the algorithms in implementation. This HIL setup consists of a CarSim PC, dSpace SCALEXIO real time simulation computer, an in-vehicle PC for running the cooperative collision avoidance algorithms, a laptop for monitoring and two DSRC on-board-unit modems for real V2V communication between the ego vehicle and the surrounding vehicles. A picture of these elements in the lab is shown in Figure 3.

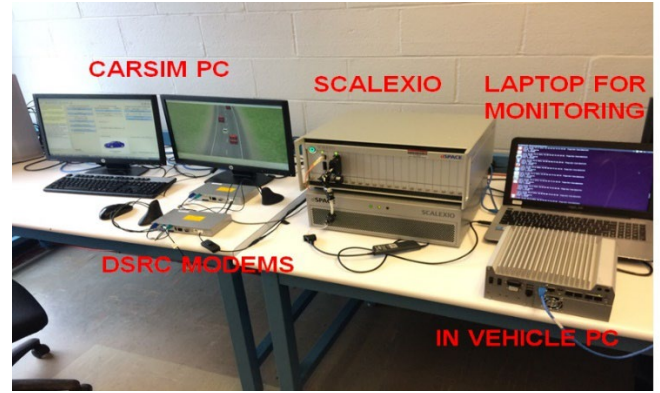

Figure 3. Picture of HIL setup.

The CarSim PC is used for creating the Simulink models and CarSim environment for the simulations. Also, during the simulation, it is used for monitoring the CarSim environment and dSpace equipment. CarSim is used as the simulation software since it has high fidelity vehicle dynamics. It is also capable of simulating the other vehicles as kinematic objects, simulating sensors and infrastructure. Moreover, vehicle dynamics parameters for the ego vehicle are determined as a result of numerous tests and experiments for validation using the real vehicle.

SCALEXIO is a real-time HIL simulation computer which runs the Simulink/CarSim models. It also has a CAN interface to communicate with other elements of the HIL setup. It sends vehicle information to the in-vehicle PC through the CAN bus where the in-vehicle PC thinks it is receiving that information from a real vehicle. SCALEXIO also sends information related to other vehicles and infrastructure to one of the DSRC modems to be broadcasted.

Controllers and decision-making are all created inside Simulink, using Simulink blocks and Stateflow for decision-making. Then, this Simulink model is converted into C code using the Simulink Coder, in order to become implementable and portable for other computers. Then, the code is implemented into the in-vehicle PC, which is a compact PC built for in-vehicle use as an electronic control unit with high processing power. This in vehicle PC runs the control algorithm in a real-time Linux environment in this setup. As a result of these control and decision-making algorithms, it sends control commands for steering, brake and throttle to the CarSim vehicle in the SCALEXIO through the CAN bus using a Kvaser CAN device. It also sends ego vehicle information to the other DSRC modem to be broadcasted and receives information from the other traffic DSRC modem. The laptop is used for connecting to the in-vehicle PC through an SSH connection and monitoring the running algorithm.

The DSRC modems emulate vehicle On Board Units (OBU). One of the modems is connected to in-vehicle PC as the ego vehicle OBU. The other modem is connected to the SCALEXIO and emulates the OBU for all of the other vehicles since it broadcast information about every vehicle other than the ego vehicle. Therefore, all of the vehicles are connected through DSRC communication. All messages are sent using the standard messages of the Society of Automotive Engineers (SAE) J2735 DSRC Message Set. The corresponding BSM (Basic Safety Message) messages carry vehicle information such as Velocity and Position published at the standard rate of 10 Hz. By using this information, the ego vehicle can make decisions for connected and autonomous driving. Communication between the HIL elements is shown in Figure 4.

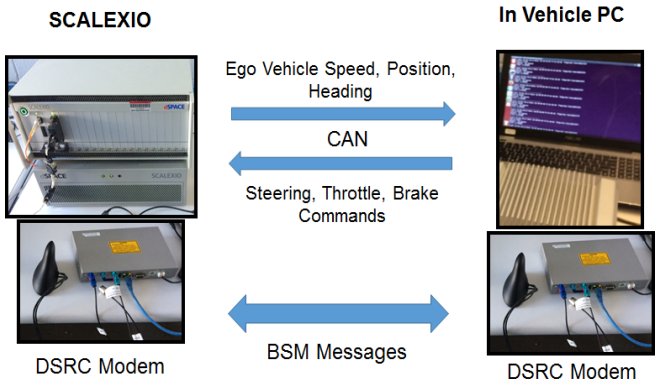

Figure 4. Communication between HIL elements.

## Simulated Scenarios

In order to simulate connected vehicle applications related to collision avoidance, 3 scenarios are selected from several high crash risk situations on the road determined by U.S.DOT. One additional scenario is selected as the 4th scenario and involves a more complex traffic situation requiring the use of the elastic band collision free path deformation and execution presented earlier in the paper. This last scenario is also modified and used with sub-scenarios to see the effectiveness of the algorithm for each case. Simulations were conducted with the HIL setup mentioned in section III, where scenario environments were created within CarSim. The ego vehicle is controlled by the in-vehicle PC and the other vehicles are simulated inside SCALEXIO and publishing information via V2V communication.

### *Electronic Emergency Brake Light Scenario*

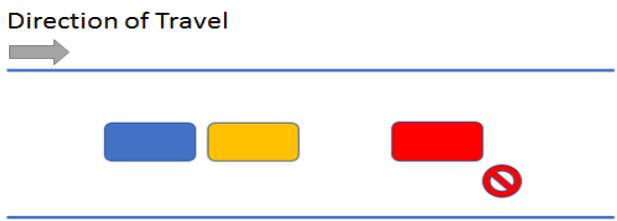

Figure 5. Illustration of the first scenario.

First scenario involves three vehicles moving on the same lane of a highway, where blue one is the ego vehicle as shown in Figure 5. The blue vehicle (ego vehicle) is tailgating the yellow vehicle ahead of it. The red vehicle brakes suddenly and comes to a quick stop. The yellow vehicle responds to the decelerating vehicle ahead by hard braking. Since, the driver of the yellow vehicle had enough time to respond, the yellow vehicle also comes to a halt without colliding with the red vehicle in front. However, since the blue vehicle had been tailgating the yellow vehicle (very small time margin for the driver to apply brakes and unaware of the sudden braking of the vehicle in front, due to lack of DSRC communication), the blue vehicle is unable to come to a halt and thus results in a rear end collision with the vehicle ahead of it. When V2V capability is added, DSRC communication allows the ego vehicle to know about the red vehicle braking in advance and be able to brake earlier to prevent the crash.

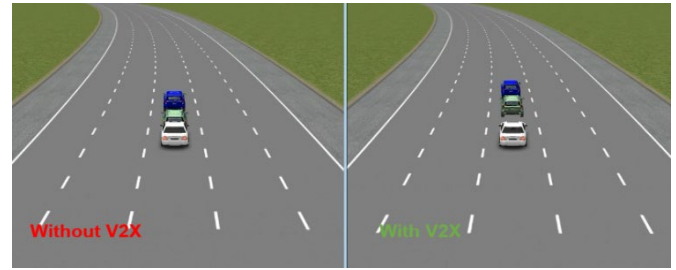

Figure 6. First scenario simulation.

As seen in Figure 6, the vehicle with V2X capability (right) stops smoothly in time and prevents the crash while the other non-communicating ego vehicle (left) crashes because it was not able to notice the vehicle braking and started to brake late.

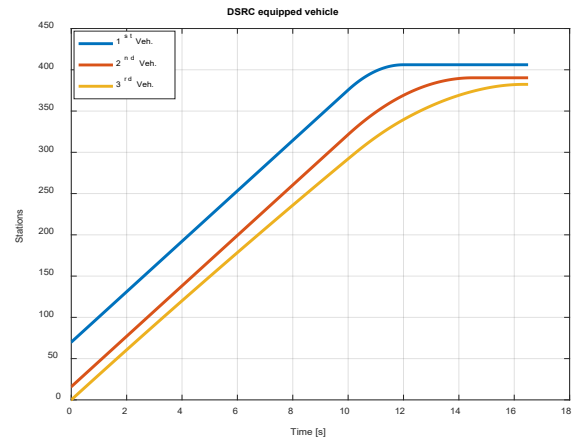

Figure 7. First scenario displacement versus time plot for all three vehicles.

As a result of the simulation for the case when the vehicles are equipped with DSRC, the displacement versus time plot for all three vehicles can be seen in Figure 7. The first vehicle immediately broadcasts the braking message to the other vehicles behind. This instantaneously activates the brake on the vehicles behind (without any delay due to the middle vehicle's reaction time). The ego vehicle comes to a stop at a safe distance from the preceding vehicle. This is evident from the Figure 7, as we can see that the position curves do not intersect one another and there is a safe space.

### *Intersection Movement Assist Scenario*

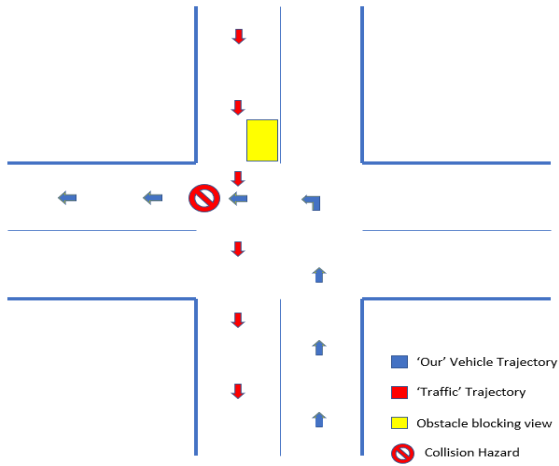

Figure 8. Illustration of the second scenario.

In the second scenario, a setting with two intersecting roads is considered. This vehicle collision scenario is shown in Figure 8. Two vehicles enter the intersection on parallel lanes but different approach directions. The trajectory of the two vehicles are defined as follows. Ego vehicle enters the intersection and decides to make a left turn at the intersection. The other vehicle enters the intersection and continues to move in the same direction. There is a stopped vehicle at the intersection, which obstructs the view of both vehicles. Both the blue and the red vehicles enter the intersection at the same time, trying to cross it, unaware of the fact that their trajectories are intersecting with one another that will result in a collision. When there is no V2V communication, since the view of the ego vehicle is obstructed by the yellow vehicle, the ego vehicle cannot detect the oncoming traffic (red vehicle). However, when there is V2V communication, the ego vehicle receives the information about the oncoming traffic that is trying to cross the intersection at high speed and waits for that vehicle to pass before attempting to make the turn.

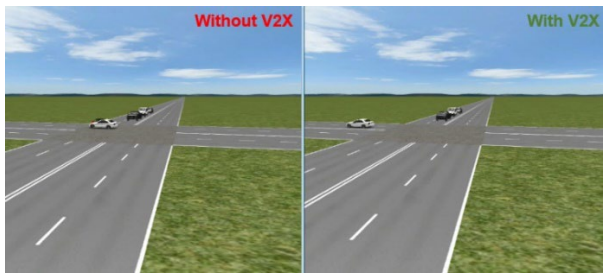

Figure 9. Second scenario simulation.

Two frames from two simulations has been shown in Figure 9. The picture on the left represents the scenario of the vehicles without the DSRC units. Without V2V communication, the ego vehicle does not have information of the oncoming traffic at the intersection (because of NLOS) and proceeds to make a turn, thus resulting in a collision at the intersection junction. The picture on the right represents the scenario of the vehicles equipped with the DSRC units. DSRC enabled vehicles inform other vehicles of their presence with V2V communication. Thus, the ego vehicle receives the broadcasted message, waits for the fast oncoming traffic to pass the intersection and only after that enters the intersection to make the turn.

## Curb Side Vehicle Alert Scenario

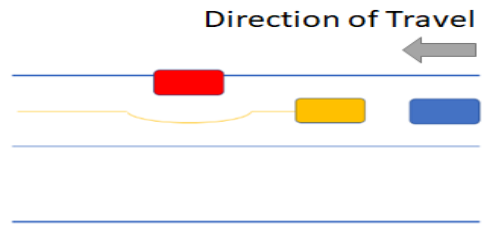

Figure 10. Illustration of the third scenario.

This scenario is illustrated in Figure 10. In this scenario, there is a vehicle parked next to the road and occupying a portion of the road (red). The vehicle directly in front of the ego vehicle (yellow) makes a late lane change around the stopped vehicle. Even though the ego (blue) vehicle cannot see the stopped vehicle, because of V2V communication, the ego vehicle is aware of the stopped vehicle and provides its decision-making algorithm with a warning ahead of time so that the controller can safely slow the vehicle and safely maneuver before reaching the stopped vehicle ahead. Without DSRC, there is very little time for the driver of the blue vehicle to respond to either brake the vehicle or to maneuver around the parked vehicle to avoid the collision with the parked vehicle and a crash occurs. With DSRC, the red vehicle communicates the information regarding its state to the nearby vehicles. The blue vehicle can either brake in a timely manner to avoid a collision or if speed is high, which is the current case, make a safe maneuver around the vehicle. While doing the autonomous drive, the ego vehicle receives the position from the red vehicle via V2V communication and applies elastic band obstacle avoidance.

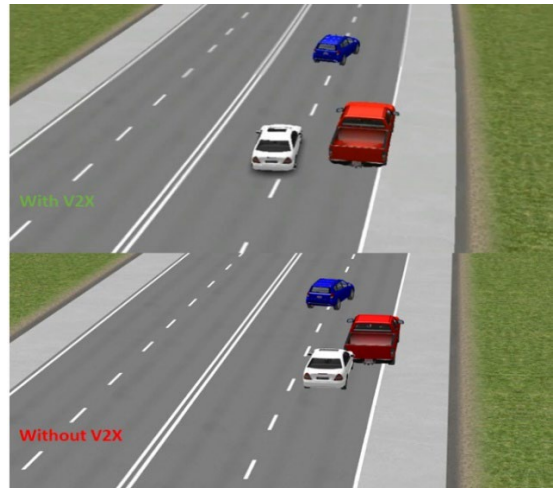

Figure 11. Third scenario simulation.

The scenario is simulated in CarSim and snapshots for with V2X and without V2X are shown in Figure 11. The red vehicle in Figure 11 represents the parked vehicle. The blue vehicle makes a sudden maneuver around the parked vehicle. Because of DSRC communication, the ego vehicle depicted in white color receives the position from the vehicle ahead of it via V2V communication and applies elastic band obstacle avoidance. But in case there is no V2X, because of the NLOS situation, ego vehicle cannot maneuver or stop in time and crash occurs.

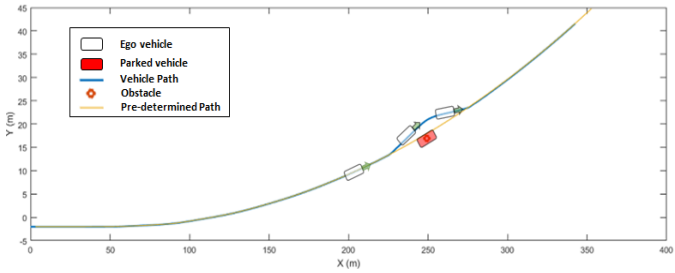

Figure 12. Third scenario ego vehicle path.

Figure 12. shows the deformed trajectory of the ego vehicle during the above scenario. The red box represents the parked vehicle along the side of the road. The yellow curve represents the original path or the curvature of the road. The blue curve shows the modified path of the ego vehicle after the application of the elastic band algorithm around the parked vehicle.

## CCA Maneuvering Scenario

The fourth scenario involves maneuvering in-between moving vehicles in traffic. It is considered for emergency situations. While performing the avoidance maneuver, the ego vehicle also considers the vehicles passing from the adjacent lane. There are three different cases which are experimented. The first case is when all the other vehicles are moving in a steady pace. For second case, there is an oncoming vehicle on the adjacent lane from behind and for the third case there is an oncoming vehicle on the adjacent lane from the opposite direction. The ego vehicle communicates with the oncoming vehicles to determine their position, velocity and direction of travel and uses this information for decision-making. Decisions are whether or not to engaging the elastic band maneuver and to increase or decrease speed. With or without V2X cases are not shown under this scenario since vehicle relies on communication to receive locations of the other vehicles and maneuver around them. Without V2X there will be no maneuvering, vehicle will stay on the lane and follow pre-determined path.

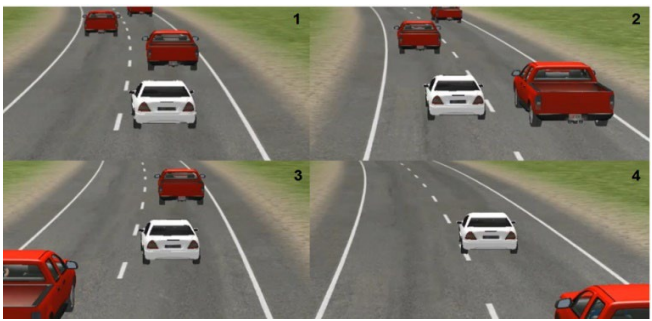

Figure 13. Fourth scenario first case simulation.

Figure 13 shows frames from simulation of the base case scenario where there are three vehicles moving on the left and the right lanes in the same direction. The ego (white) vehicle is behind all the remote (red) vehicles and tries to get ahead of the slow-moving traffic. The ego vehicle then applies the elastic band algorithm automatically in-between the three slow moving vehicles to get ahead of them. After the simulation, the vehicle path is plotted on Figure 14 and positions of the remote vehicles and ego vehicle are illustrated according to the frame numbers on Figure 13.

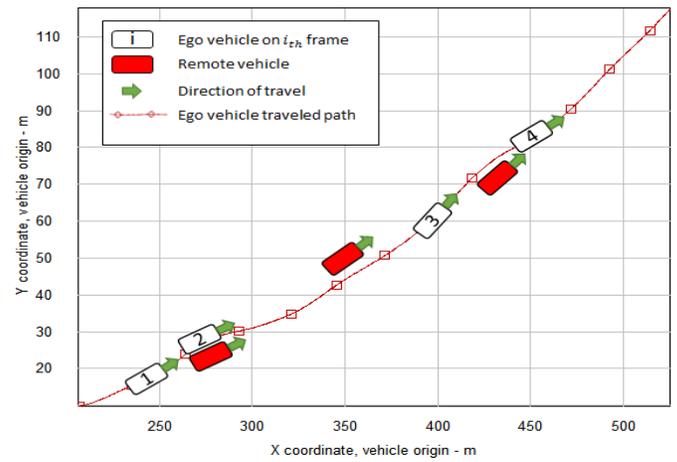

Figure 14. Fourth scenario first case vehicle path and scenario illustration.

The collision free trajectory of the ego vehicle and the three obstacle vehicles during this maneuver are shown in Figure 14. The ego vehicle maneuvers safely between the other three remote vehicles while travelling along the pre-defined path. It is also important to note that all vehicles in this illustration as well as following illustrations are moving objects, but remote vehicles are shown only for a single frame to have a clear and more readable plot. Maneuvering behavior along the path can also be tuned for different types of responses. Two different avoidance paths for fourth scenario simulation is shown in Figure 15. If parameters are tuned for faster response, vehicle will maneuver faster, providing quicker lane change and centering but less comfort. The opposite is tuning for smoother response which maneuvers slower but provides more comfort.

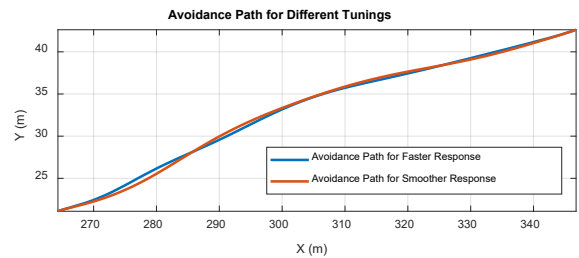

Figure 15. Fourth scenario first case simulation.

Figure 16 shows a slight modification to the base case mentioned earlier. The scenario is defined as follows. There is a vehicle moving on the right lane ahead of the ego vehicle. The ego vehicle tries to move in front of this vehicle by making a maneuver around the vehicle in front. There is another vehicle approaching at higher speed in the same direction on the left lane. If the ego vehicle makes a maneuver without knowledge of the vehicle on the left lane, there would be a collision. When the vehicles are equipped with DSRC communication, the ego vehicle gets information about the speeding vehicle on the left lane, thus instead of maneuvering, the ego vehicle uses decision-making calculations mentioned previously, to adapt to the speed to the preceding vehicle, allowing the speeding vehicle on the left lane to pass and then uses the elastic band algorithm to make a safe maneuver around the vehicle on the right lane.

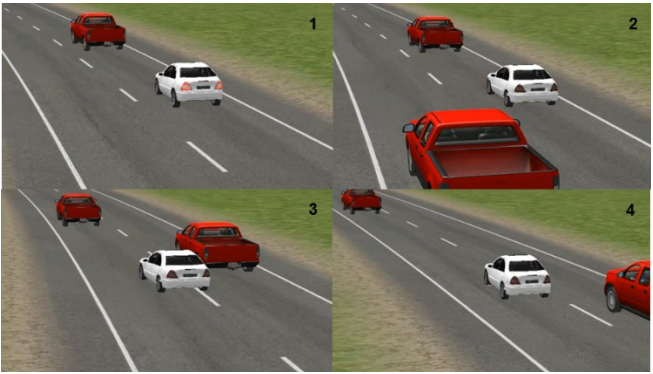

Figure 16. Fourth scenario second case simulation.

For the fourth scenario second case, vehicle path and illustration of the locations of ego vehicle, remote vehicles and approaching vehicle are shown in Figure 17. Frame numbers are placed according to the frame numbers shown in Figure 16. By looking at the vehicles shown in the figure, it can be seen that the ego vehicle waits within the first and second frames and follows the pre-defined path since there is a vehicle approaching from behind on the adjacent lane. Only around the third frame, after the adjacent lane vehicle passes, does the ego vehicle start maneuvering around the second remote vehicle. The second remote vehicle and the elastic band based maneuvering are also illustrated in Figure 17.

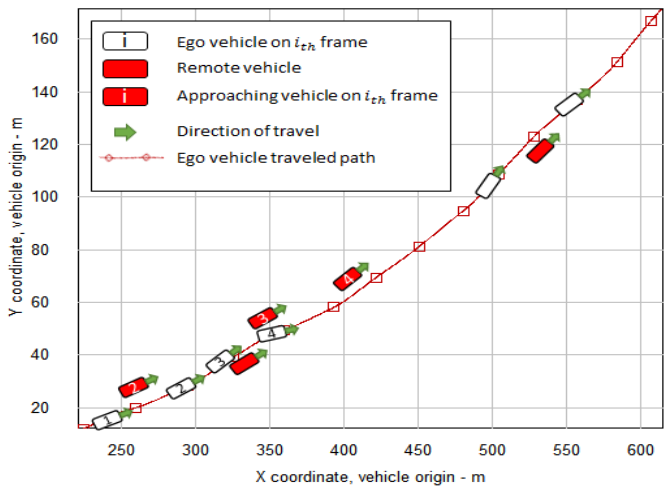

Figure 17. Fourth scenario second case vehicle path and scenario illustration.

Figure 18 represents another modification to the base case mentioned earlier. The scenario is defined as follows. There is a vehicle moving on the right lane ahead of the ego vehicle. The ego vehicle tries to move in front of this vehicle by making a maneuver around the vehicle in front. There is another vehicle approaching at higher speed in the opposite direction on the left lane. If the ego vehicle makes a maneuver without knowledge of the vehicle on the left lane, there would be a collision. When the vehicles are equipped with DSRC communication, the ego vehicle gets information about the oncoming traffic on the other lane, thus instead of maneuvering, the ego vehicle uses decision-making mentioned earlier to adapt its speed to the preceding vehicle, lets the oncoming traffic pass, and then makes a maneuver around the preceding vehicle when it is safe to do so (i.e., there is no more oncoming traffic or the oncoming traffic is at a very large distance).

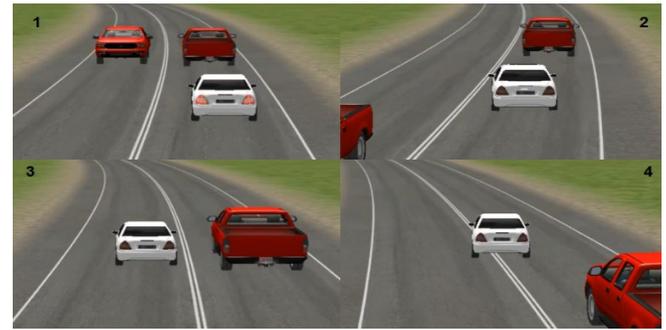

Figure 18. Fourth scenario third case simulation snapshots.

For the last case in this scenario, the vehicle path and illustration of the locations of the ego vehicle, preceding vehicle and approaching vehicle are shown in Figure 19. Frame numbers are placed according to the frame numbers shown in Figure 18. When the ego vehicle, approaching vehicle and preceding vehicle locations are analyzed in Figure 19, it can be seen that the ego vehicle waits for the approaching vehicle from the opposite direction until the second frame. Around the second frame, the approaching vehicle passes and the ego vehicle starts maneuvering around the preceding vehicle. First maneuvering around the preceding vehicle is also illustrated but not shown in screenshots, since the behavior is very similar to the first case.

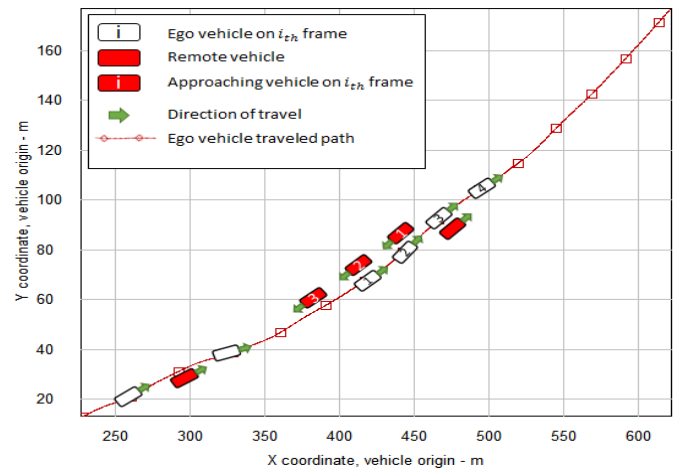

Figure 19. Fourth scenario third case vehicle path and scenario illustration.

## Conclusion

To emphasize and explore the opportunities that safety related V2V applications create, a communication based decision making and collision avoidance method is introduced that uses elastic band theory for modification of the path when necessary. Three different high crash risk NLOS scenarios were selected and successfully simulated as well as compared to simulated cases without V2X. Where a fourth scenario was added to assess the behavior in more complex situations that involves maneuvering around multiple vehicles. Furthermore, HIL simulations were conducted in real-time to evaluate the algorithm and decision-making performance inside the CarSim environment with a high fidelity, validated vehicle model and results are shown. Future work will involve real world experiments with our simulation vehicle platforms. Future work can also involve integrating or combining some of the approaches in this paper with other control, AV, CV, ADAS, automotive control and other topics like those in references [16-77] and others in the future.

## Contact Information

Sukru Yaren Gelbal    gelbal.1@osu.edu

Automated Driving Laboratory

1320 Kinnear Road, Columbus, OH, 43212

## Acknowledgments

This paper is based upon work supported by the Ohio State University Center for Automotive Research Membership Exploratory Research Project titled: Cooperative Collision Free Path Planning and Collision Avoidance for Autonomous Driving. The support of the CAR Membership Consortium is gratefully acknowledged.

## Definitions/Abbreviations

| | |
|---|---|
| **CV** | Connected Vehicle |
| **V2V** | Vehicle to Vehicle |
| **V2I** | Vehicle to Infrastructure |
| **V2P** | Vehicle to Pedestrian |
| **NLOS** | No Line of Sight |
| **CCA** | Cooperative Collision Avoidance |
| **V2X** | Vehicle to Everything |
| **CAV** | Connected Autonomous Vehicle |
| **HIL** | Hardware in the Loop |
| **DOT** | Department of Transportation |
| **DSRC** | Dedicated Short-Range Communication |
| **OBU** | On Board Unit |
| **GPS** | Global Positioning Systems |
| **CAN** | Controller Area Network |
| **SSH** | Secure Shell |
| **SAE** | Society of Automotive Engineers |
| **BSM** | Basic Safety Message |